\documentclass[letterpaper, 10 pt, conference]{ieeeconf}  

\IEEEoverridecommandlockouts                              

\usepackage[table]{xcolor}
\usepackage{wrapfig}
\usepackage{pifont}
\usepackage{array}
\usepackage{hyperref}
\usepackage{url}

\usepackage{booktabs,makecell}
\usepackage{wrapfig}

\usepackage{paracol}
\usepackage{amsmath,amssymb,algorithm,algorithmic}

\usepackage{overpic}
\usepackage{adjustbox}
\usepackage{colortbl}
\usepackage{multirow}
\usepackage{booktabs}
\usepackage{tabularx} 
\usepackage{booktabs,multirow,makecell,array,graphicx}
\definecolor{eccvblue}{rgb}{0.21,0.49,0.74}
\usepackage{threeparttable}
\usepackage{siunitx}
\usepackage{wrapfig}
\usepackage{array}
\usepackage{longtable}
\usepackage{calc}      
\usepackage{caption}   
\usepackage{tikz}
\usetikzlibrary{positioning,arrows.meta}
\usepackage{caption}
\usepackage{subcaption}      
\usepackage{amsmath}
\usepackage{graphicx}
\usepackage{booktabs}
\definecolor{cellgray}{RGB}{200,200,200}
\usepackage{array}
\usepackage{booktabs}
\usepackage{multirow}
\usepackage{makecell}
\usepackage{hyperref}
\newcolumntype{L}[1]{>{\raggedright\arraybackslash}p{#1}}
\usepackage{cuted}
\usepackage{capt-of}
\usepackage[accsupp]{axessibility}  

\usepackage{orcidlink}
\usepackage{hyperref}
\usepackage{wrapfig}
\usepackage[nameinlink,capitalize]{cleveref}
\usepackage{xspace}
\usepackage{placeins}
\title{\LARGE \bf
\raisebox{-0.15\height}{\includegraphics[height=0.9em]{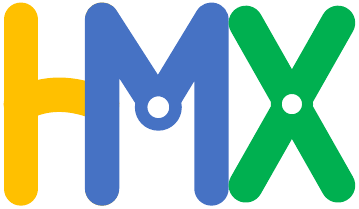}}
\textbf{\textit{HandMimicX}}: Learning Human-Like Dexterous Manipulation via an Object-Centric Policy from Egocentric Videos
}

\author{
\begin{tabular}{@{}c@{}}
{\normalfont\mdseries
Zihan Gao$^{1,2,3,*}$ \quad
Yufei Liu$^{2,4,*}$ \quad
Yuhong Zhang$^{1,2,\ddagger}$ \quad
Zhen Yu$^{2}$ \quad
Xinhao Huang$^{1,2}$}\\
{\normalfont\mdseries
Zhinan You$^{2}$ \quad
Kai W. Ho$^{5}$ \quad
Zipeng Ye$^{3,\dagger}$ \quad
Peng Jiao$^{1,2}$ \quad
Haoqian Wang$^{1,\dagger}$}\\[0.3em]
{\normalfont\bfseries
$^{1}$Tsinghua University \quad
$^{2}$TsingTec \quad
$^{3}$Meituan}\\
{\normalfont\bfseries
$^{4}$The Chinese University of Hong Kong, Shenzhen \quad
$^{5}$University of Southern California}\\[0.6em]
\end{tabular}%
\vspace{-1em}
\thanks{%
\hspace*{-1.8em}%
\makebox[\linewidth][c]{%
$^{*}$Co-first authors. \quad
$^{\ddagger}$Project lead. \quad
$^{\dagger}$Corresponding authors.%
}\protect\newline
\makebox[\linewidth][c]{%
Email: \{gaozihanthu,\,liuyufei200003\}@gmail.com
}}
 }

\begin{document}

\maketitle
\thispagestyle{empty}
\pagestyle{empty}

\begin{strip}
\vspace{-5.0em}
    \centering
    \includegraphics[width=\textwidth]{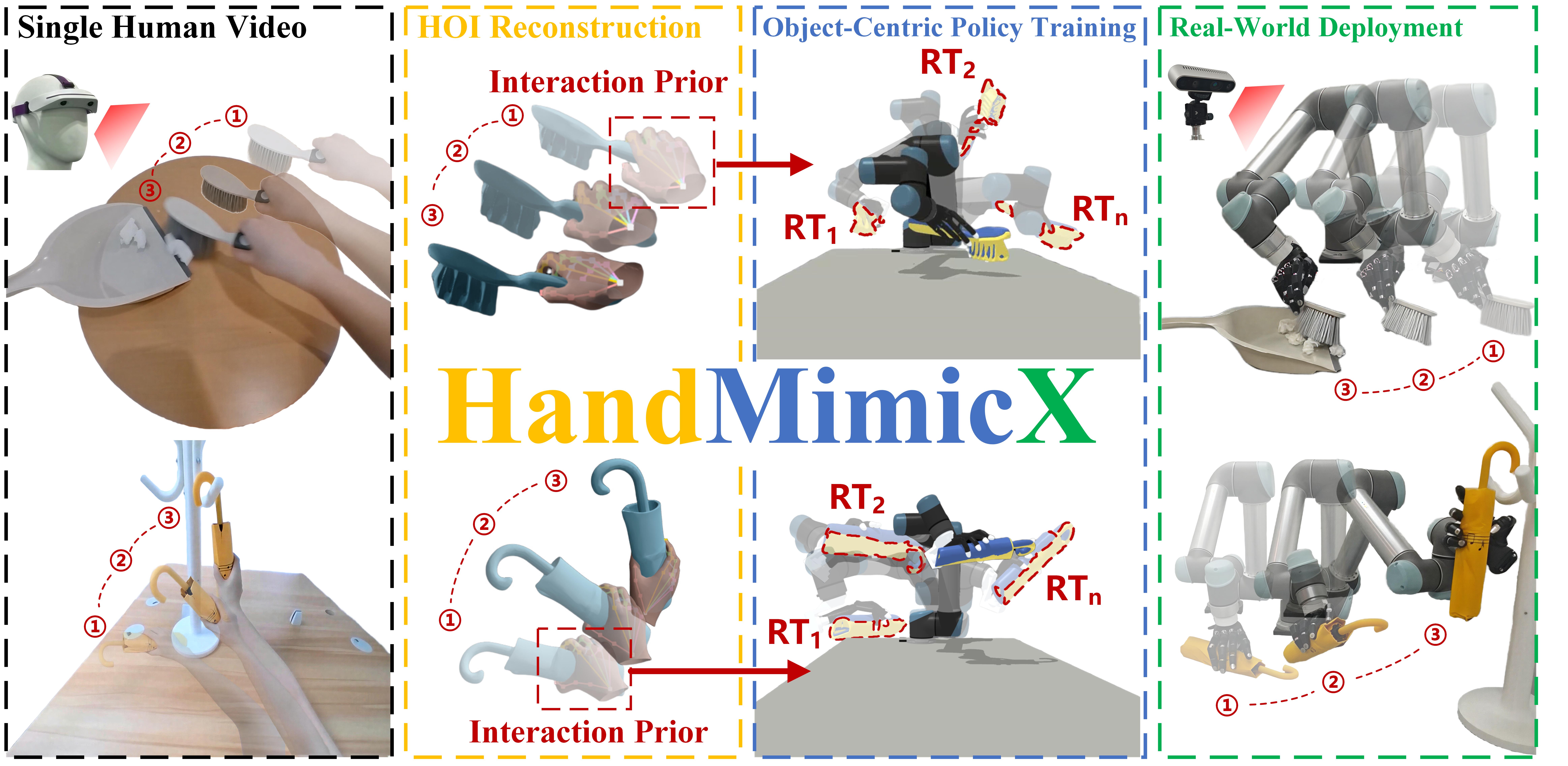}
    \captionof{figure}{
    \textbf{\textit{HandMimicX}}: 
We present an object-centric framework for learning dexterous manipulation skills from a single egocentric demonstration, leveraging simulation-based reinforcement learning and sim-to-real policy transfer. The framework consists of three stages. First, we reconstruct the demonstrated hand–object interaction from the input video, recovering the hand–object motion trajectory, object geometry, and task-relevant interaction priors. Second, using the reconstructed object mesh and interaction priors, we train an object-centric tracking policy to manipulate the object to spatially randomized target object poses, denoted as $\mathrm{RT}_n$. By optimizing the policy across diverse target poses, the robot acquires spatial generalization beyond the original human demonstration. Finally, the learned policy is transferred to the real robot and executed using observations from an RGB-D camera, enabling dexterous manipulation in the real-world environment.
    }
    \label{fig:teaser}
    \vspace{-0.5em}
\end{strip}

\begin{abstract}
Learning dexterous manipulation from human videos remains an open challenge, particularly in the context of transferring a single demonstration into a spatially generalizable robotic skill and maintaining physically plausible human-like interactions. Unlike prior works that emphasize hand-motion imitation,
we introduce an object-centric framework for acquiring dexterous skills from egocentric human RGB-D videos. Within our framework, we formulate the reinforcement learning (RL) objective in terms of object motion to decouple the underlying skill from the specific hand trajectory and impose constraints on hand–object interaction states to promote more anatomically natural hand configurations. Extensive experiments in simulation and on real robots demonstrate that our method enables efficient learning from human videos and generalizable skill transfer across novel spatial configurations. Project page:
\url{https://anonymous.4open.science/w/handmimicx/index.html}.

\end{abstract}

\section{Introduction}
\label{sec:intro}

Recent advances in embodied intelligence have demonstrated the potential of learning robot manipulation policies from human videos~\cite{zheng2026egoscale,luo2025being,mandikal2022dexvip,qin2022dexmv}. However, directly transferring
human demonstrations to robots remains challenging due to the gap between human motion and
robot embodiments, as well as the discrepancy between visual observations and physical execution~\cite{Yuan2025HERMESHE,Zhang2026UniDexAR,Han2026Video2Sim2RealFA}. 

Reinforcement learning (RL) provides a powerful mechanism to bridge this gap by optimizing interaction
behaviors under physical constraints in simulation. Under this paradigm,
 human demonstrations can be transformed into physically feasible robot manipulation
policies. However, acquiring a robust and generalizable skill from a single human demonstration remains challenging. Existing RL approaches largely rely on hand motion as the primary supervision signal, learning residual actions to refine the reference motion~\cite{Yuan2025HERMESHE,Li2025ManipTransED,Liu2025DexTrackTG,Feng2026AMR}. While this paradigm enables precise motion imitation, it relies heavily on accurately reconstructed hand–object trajectories, making performance sensitive to even minor misalignments and favoring kinematic reproduction over the acquisition of transferable skills. By contrast, object-centric policies~\cite{Kuang2026Dex4DTP,Kedia2026SimToolRealAO} prioritize desired object states, shifting the learning objective from replicating a demonstration to capturing what the task is meant to accomplish. 
Nevertheless, their disproportionate emphasis on object-level task success often leads to unrealistic hand–object interactions, including physically implausible contacts and unnatural finger configurations. 

In light of these challenges, we propose \textbf{HandMimicX}, a novel object-centric reinforcement learning framework for acquiring an underlying manipulation skill from a single egocentric human video. The ``X'' indicates moving beyond trajectory mimicry toward generalizable skill acquisition. To bridge egocentric human videos and dexterous control, we jointly reconstruct hand-object interactions (HOIs). However, camera motion degrades reconstruction accuracy, particularly in terms of hand–object alignment. We therefore refine the reconstructed hand translation using depth to improve spatial consistency. To transcend the trajectory constraints inherited from the human demonstration video, we condense the full-sequence reconstructions into object geometry and interaction priors to retain the most critical task-relevant elements. Using the reconstructed geometry, we train an RL policy to manipulate the object to arbitrary 6-DoF target poses, while the interaction prior regularizes the resulting hand–object interactions toward physically plausible configurations. For real-world deployment, we use DAgger~\cite{ross2011dagger} to distill the policy and reduce its reliance on privileged information. Ultimately, our framework transforms a single human demonstration video into a transferable dexterous skill that generalizes beyond the demonstrated spatial configuration. \noindent\textbf{In summary, our contributions are as follows:}
\begin{itemize}
    \item We propose \textbf{HandMimicX}, a novel object-centric RL framework that learns a generalizable dexterous manipulation skill from a single human video.
    \item We build a joint 4D HOI reconstruction pipeline that recovers spatially consistent hand–object motion from dynamic egocentric views.
    \item We enhance object-centric RL with an extracted interaction prior, regularizing policy learning toward physically plausible and anthropomorphic dexterous manipulation.
\end{itemize}

\section{Related Work}
\label{sec:related_work}

\subsection{Human--Object Interaction Reconstruction}
\label{sec:related_hoi_reconstruction}

Recovering human--object interactions (HOIs) from visual observations provides the geometric interface between human videos and downstream robot learning~\cite{zhang2026robowheel}. HOLD~\cite{fan2024hold} leverages a compositional articulated implicit
model to jointly reconstruct articulated hands and objects using compositional SDF and contact constraints. HO-NeRF~\cite{qu2023novel} separately learns
shape and appearance prior knowledge of hands and
objects with a neural representation and proposes a pose estimation system for hand-object interaction
from sparse views. These methods, however, rely on either scene-specific optimization or pre-built hand models, limiting their scalability to in-the-wild videos. Another line of work, exemplified by HaMeR~\cite{pavlakos2024reconstructing}, scales up a ViT-based backbone for 3D hand reconstruction, yielding accurate and robust hand mesh recovery, while HaWoR~\cite{zhang2025hawor} and Dyn-HaMR~\cite{yu2025dyn} couple camera tracking with hand-motion estimation for global 4D hand recovery from unconstrained videos. Recent studies~\cite{zhang2026robowheel,chen2026dexterous,ren2026c2dex} have developed more generalizable integrated frameworks for reconstructing overall hand--object motion. Nevertheless, these approaches remain susceptible to cumulative errors, particularly in hand--object alignment, compromising downstream manipulation.

\subsection{Learning Robot Manipulation from Human Videos}
\label{sec:related_human_video_learning}

Exploiting human videos to train robotic models has been
actively studied in recent years. Recently,~\cite{zheng2026egoscale,luo2025being,li2025scalable,bi2026h} have utilized human video for pretraining robotic manipulation models; for
instance, EgoScale~\cite{zheng2026egoscale} uncovers a
clear log-linear relationship between data scale and hand-action prediction loss and demonstrates that this loss
strongly correlates with real-robot dexterous manipulation performance. By leveraging large-scale human video corpora~\cite{hoque2026egodex,chao2021dexycb,banerjee2025hot3d,zhan2024oakink2,Kwon2021H2OTH}, these works envision a future in which robots acquire diverse skills from abundant human demonstrations. In parallel, advances in simulation~\cite{makoviychuk2021isaac,todorov2012mujoco} have established real-to-sim-to-real as an alternative paradigm for robot learning from videos~\cite{mandikal2022dexvip,qin2022dexmv,Chen2024ViViDexLV,Xu2026DemoBotEL}, in which policies are optimized via RL to reproduce motions demonstrated in human videos. 
Our proposed \textbf{HandMimicX}, instead, emphasizes learning a transferable skill via RL from a single human demonstration video, moving beyond mere imitation of the demonstrated trajectories.

\subsection{Object-Centric Reinforcement Learning}
\label{sec:related_object_centric_rl}
Unlike residual RL~\cite{Li2025ManipTransED,Xu2026DemoBotEL,Ye2025Gen2RealTD}, which remains anchored to a reference trajectory, object-centric RL~\cite{Zhou2023LearningHA,Qin2022DexPointGP} directly optimizes task-relevant object-state transitions. Dex4D~\cite{Kuang2026Dex4DTP} learns a generalizable task-agnostic grasping policy, while X-Sim~\cite{Dan2025XSimCL} trains a reinforcement
learning (RL) policy in photorealistic simulations reconstructed from RGB-D human videos. To simplify RL for robotic hand control, Human2Sim2Robot~\cite{Lum2025CrossingTH} combines geometric fabric control with a PCA-based action space, but this formulation can lead to degenerate hand configurations. These prior methods either target grippers or neglect the naturalness and anthropomorphism of hand configurations during grasping. By contrast, we transfer physically grounded hand–object interaction patterns from human videos to promote feasible dexterous manipulation.

\section{Method}
\label{sec:method}

\begin{figure*}[!t]
    \centering
    \includegraphics[width=\textwidth]{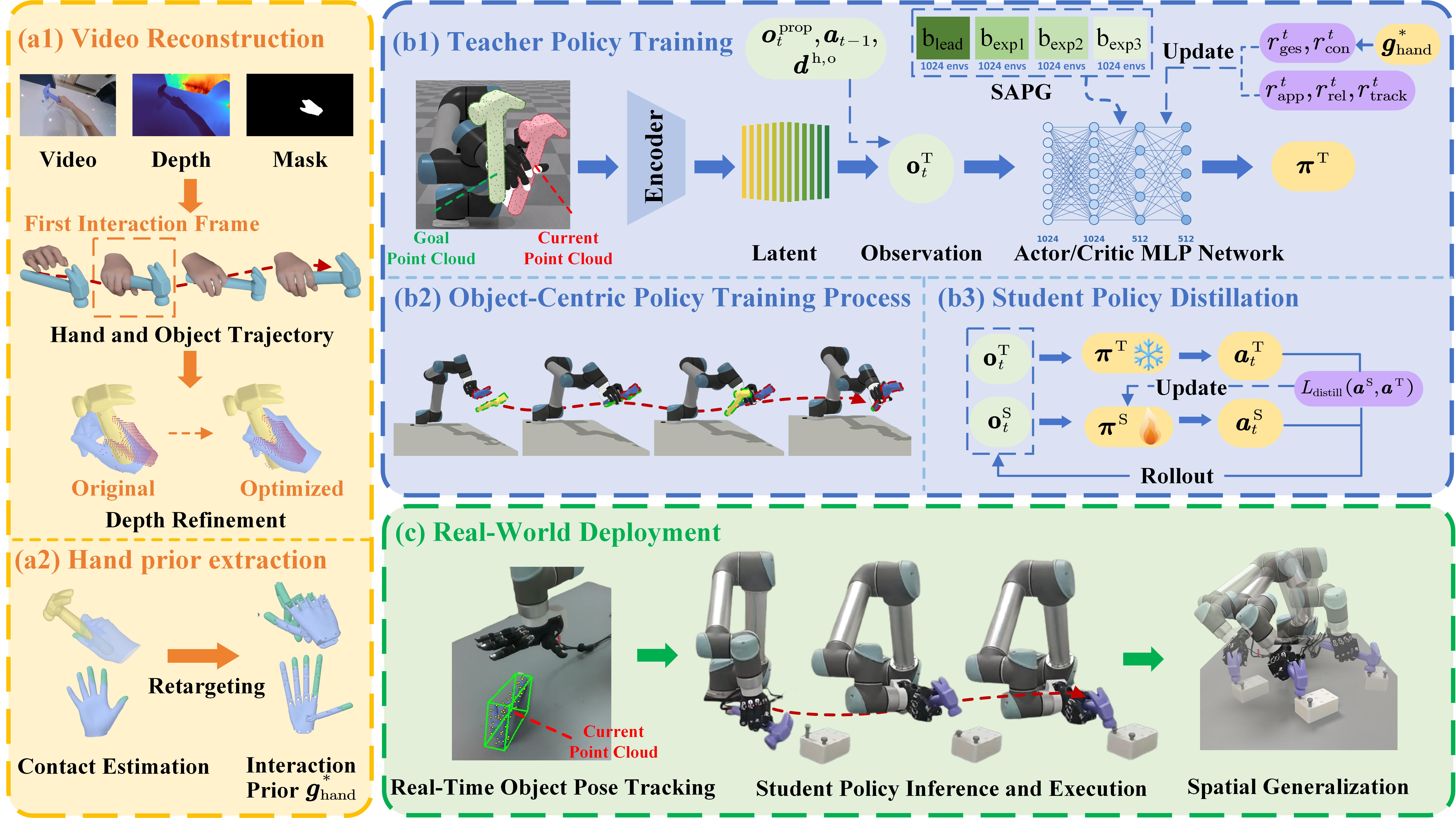}
    \caption{Overview of \textbf{HandMimicX}. (a1) Given an RGB-D egocentric video, we reconstruct the manipulated object and hand motion, and then align both reconstructions with the depth observations. (a2) We extract the hand-object interaction state at the first contact frame as the interaction prior $\boldsymbol{g}_{\mathrm{hand}}^{*}$ for policy learning. (b1) We train an object-centric teacher policy $\boldsymbol{\pi}^{\mathrm T}$ using PPO with SAPG.  A correspondence-aware encoder maps the current and goal object point clouds to a geometric latent representation, which is concatenated with $\boldsymbol{o}_t^{\mathrm{prop}}$, $\boldsymbol{a}_{t-1}$ and $\boldsymbol{d}^{\text{h,o}}$ to form the observation $\boldsymbol{o}_t^{\mathrm{T}}$ and then fed into the actor and critic MLPs. 
    (b2) During training, the teacher $\boldsymbol{\pi}^{\mathrm T}$ learns to manipulate the object and track arbitrary goal poses. (b3) We use DAgger to distill the teacher $\boldsymbol{\pi}^{\mathrm T}$ into a student $\boldsymbol{\pi}^{\mathrm S}$ under partial observations $\boldsymbol{o}_t^{\mathrm S}$.
    Finally, (c) we deploy the distilled student policy on the real robot to execute diverse trajectories, extending beyond the original video demonstration.
    }
    \label{fig:pipeline}
    \vspace{-1em}
\end{figure*}

\subsection{Overview}
An overview of our framework is shown in Fig.~\ref{fig:pipeline}. Given a single egocentric RGB-D video, we reconstruct a depth-refined HOI trajectory to extract the metric-scale object mesh and the interaction prior. Next, we train an object-centric policy under the constraints of the prior in simulation. For deployment, we use DAgger~\cite{ross2011dagger} to distill the privileged policy trained in simulation into a student policy conditioned on partial observations available in the real world.

\subsection{Egocentric HOI Reconstruction}
\label{sec:interaction prior extraction}
\subsubsection{Problem Setup}
Given a dynamic egocentric video sequence ${I_t}_{t=1}^{T}$, we formulate the training objective based on the following physical state representations: the geometry of the manipulated object and interaction priors capturing hand configuration, hand–object relative pose, and contact.

\subsubsection{Joint HOI Reconstruction with Depth Refinement}

\begin{figure}[!h]
    \centering
    \includegraphics[width=\linewidth]{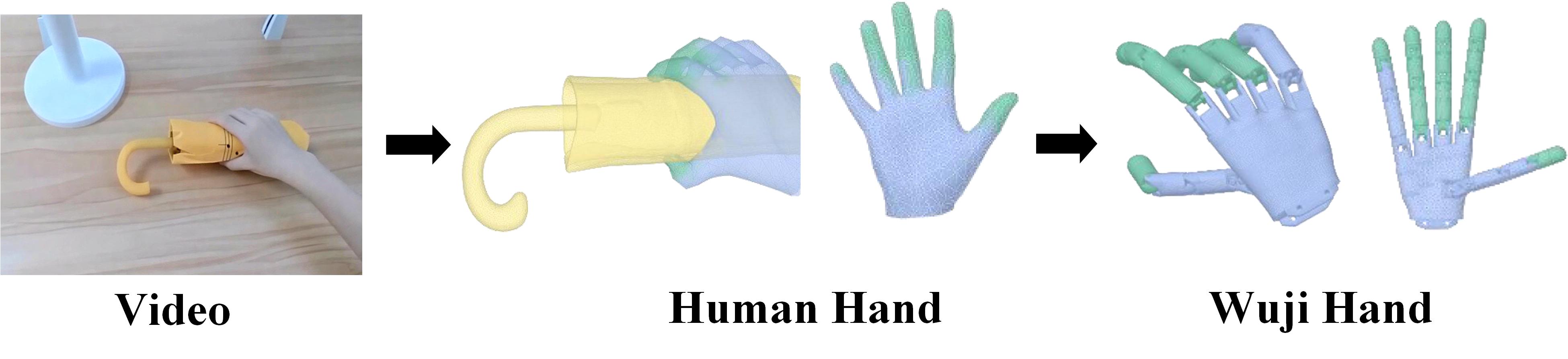}
    \caption{\textbf{Visualization of contact and hand pose retargeting.} We use HACO to estimate contact regions on the MANO hand mesh and transfer them to the robotic hand for contact retargeting. Hand pose retargeting is performed using~\cite{Qin2023AnyTeleopAG}.}
    \label{fig:contact_retarget}
    \vspace{-1.5em} 
\end{figure}

We aim to recover spatially and temporally consistent hand--object interactions from dynamic egocentric observations. Specifically, we use SAM3D~\cite{chen2026sam3d} to recover the metric-scale object mesh, HaWoR~\cite{zhang2025hawor} to estimate hand poses in the world coordinate frame, and HACO~\cite{jung2025haco} to detect hand--object contact and identify the corresponding contact regions, as shown in Fig.~\ref{fig:contact_retarget}. The interaction prior is then defined by the hand pose, relative hand--object position, and region-level contact at contact onset.  We additionally reconstruct the demonstrated object trajectory with FoundationPose~\cite{Wen2023FoundationPoseU6} throughout the video. Although this trajectory is not directly used for policy training, it provides a reference for real-world execution and serves as a basis for generating additional object motion trajectories.

However, a key challenge is that independently reconstructing the hand and object under continuous ego-camera motion can introduce scale drift and spatial misalignment, leading to inconsistent interaction trajectories. 
To mitigate the inconsistency between reconstructed hand and object trajectories, we further refine the global hand translation using metric depth observations. 
For each interaction frame $t$, we back-project valid depth pixels within the hand mask into a world-space point cloud and randomly sample 10 points to form a sparse set of depth anchors $\mathcal{P}_t$. 
We then identify the MANO mesh vertices visible to the camera and obtain their world-space coordinates, denoted by $\mathcal{V}_t$. 
The global hand translation is refined by aligning the visible hand surface with the depth observations:
\begin{equation}
    \Delta \mathbf{t}_t^{*}
    =
    \arg\min_{\Delta \mathbf{t}}
    \sum_{\mathbf{v}\in\mathcal{V}_t}
    \rho\!\left(
        \min_{\mathbf{p}\in\mathcal{P}_t}
        \left\|\mathbf{v}+\Delta\mathbf{t}-\mathbf{p}\right\|_2^2
    \right).
\end{equation}
We adopt the Huber loss $\rho(\cdot)$ to reduce sensitivity to depth outliers. 
The refined hand translation is
$\mathbf{t}_t=\mathbf{t}_t^{0}+\Delta\mathbf{t}_t^{*}$.
By anchoring the hand reconstruction to metric depth observations, this refinement alleviates scale drift and hand--object misalignment caused by ego-camera motion, substantially reducing interpenetration and improving the spatial and temporal consistency of the recovered interaction.

\subsection{Object-Centric Policy for Learning from Videos}
\label{sec:RL policy}
\subsubsection{Object-Centric Policy Design}
Our policy $\boldsymbol{\pi}$ employs an actor--critic architecture to jointly control the 6-DoF UR5 arm and the 20-DoF Wuji hand. At each time step $t$, the actor receives an observation $\boldsymbol{o}_t$ consisting of the
current and target object point clouds, robot joint positions and velocities $\boldsymbol{o}_{t}^{\text{prop}}$, the previous action $\boldsymbol{a}_{t-1}$, and distances $\boldsymbol{d}^{\text{h,o}}$ from the fingertips to their nearest object points. We employ a correspondence-aware encoder to capture geometric relationships between the current and target object point clouds while preserving pointwise correspondences, yielding a compact and expressive latent representation of their relative pose. The remaining observations are concatenated and passed through an MLP encoder to obtain a proprioceptive embedding.
All embeddings are jointly fed into the actor to produce a 26-dimensional action vector. The predicted action is executed in the simulation environment to interact with the object and receive the corresponding reward. Our critic estimates the expected cumulative future return from the current state under the actor policy, providing a value signal to guide policy optimization.
Throughout the process, the interaction prior shapes the reward by measuring the deviation from the reference hand pose, encouraging the policy to maintain physically plausible hand configurations. Additional privileged information, including contact and collision signals, is used to assess policy performance.

\paragraph{Point Correspondence Learning}
Motivated by~\cite{fu2025cordvip}, we use point correspondences to capture the geometric relationship between the current and goal object states. Let
$\bar{\mathbf{X}}=\{\bar{\mathbf{x}}_i\}_{i=1}^{N}$
denote a set of $N = 128$ points sampled from the object mesh. 
A shared PointNet~\cite{Qi2016PointNetDL} encoder $\phi_{\theta}$ extracts pointwise embeddings from the current and goal point clouds:
$\mathbf{H}^{s}_{t}=\phi_{\theta}(\mathbf{X}^{s}_{t})$, where
$s\in\{c,g\}$, with $c$ and $g$ denoting the current and goal object point clouds, respectively.
To capture the correspondence between the current and goal object states, we apply bidirectional cross-attention:
\begin{equation}
\begin{aligned}
\widetilde{\mathbf{H}}^{c}_{t}
&= g_{\theta}(\mathbf{H}^{c}_{t},\mathbf{H}^{g}_{t})
 + \mathbf{H}^{c}_{t}, \\
\widetilde{\mathbf{H}}^{g}_{t}
&= g_{\theta}(\mathbf{H}^{g}_{t},\mathbf{H}^{c}_{t})
 + \mathbf{H}^{g}_{t}.
\end{aligned}
\end{equation}
Here, $\mathbf{H}^{c}_{t}$ and $\mathbf{H}^{g}_{t}$ denote the pointwise features of the current and goal object point clouds, respectively, and $g_{\theta}$ denotes the transformer-based cross-attention module. The resulting features $\widetilde{\mathbf{H}}^{c}_{t}$ and $\widetilde{\mathbf{H}}^{g}_{t}$ mutually encode the two geometries, capturing their correspondence and producing interaction-aware representations for fusion.
\paragraph{Reward Functions}
The total reward is designed to encourage the robotic hand to manipulate the object toward the target poses while promoting plausible interactions:
\begin{equation}
\begin{aligned}
r_t ={}&
w_{\mathrm{app}}r_{\mathrm{app}}
+
w_{\mathrm{con}}r_{\mathrm{con}}
\\
&+
w_{\mathrm{ges}}r_{\mathrm{ges}}
+
w_{\mathrm{rel}}r_{\mathrm{rel}}
+
w_{\mathrm{track}}r_{\mathrm{track}}.
\end{aligned}
\end{equation}
The approach reward $r_{\mathrm{app}}$ encourages the robotic hand to approach the object by reducing the finger--object distance $\boldsymbol{d}^{\text{h,o}}$. The contact reward $r_{\mathrm{con}}$ encourages the target
regions of the robotic hand to make contact with the object.

To identify the target contact regions, we first use HACO to estimate contacts on the human hand and then retarget them to the robotic hand, as illustrated in Fig.~\ref{fig:contact_retarget}. Specifically, HACO predicts a binary contact label $h_t(\boldsymbol{v})\in\{0,1\}$ for each MANO vertex $\boldsymbol{v}$. We aggregate these vertex-level predictions into region-level contact estimates using a predefined anatomical segmentation. These contact estimates are then transferred to the corresponding semantic regions on the Wuji hand, each associated with a set of candidate robot links. The resulting set of target contact regions, denoted by $\mathcal{C}$, is used to define the contact reward:
\begin{equation}
r_{\mathrm{con}} =
\sum_{i \in \mathcal{C}}
I\left(d^{\text{h,o}}_{i} < \epsilon\right),
\end{equation}
where $d^{\text{h,o}}_{i}$ denotes the distance between the $i$-th robotic hand region and the object, $\epsilon$ is the contact threshold, and $I(\cdot)$ is the indicator function. 
The relative-pose reward $r_{\mathrm{rel}}$ preserves the demonstrated
hand--object configuration by penalizing deviations in both
relative translation $\mathbf{P}_{\mathrm{rel}}^{t}
-\hat{\mathbf{P}}_{\mathrm{rel}}^{t}$ and rotation $\mathbf{R}_{\mathrm{rel}}^{t}
\ominus\hat{\mathbf{R}}_{\mathrm{rel}}^{t}$,
where $\mathbf{P}_{\mathrm{rel}}^{t}$ and
$\mathbf{R}_{\mathrm{rel}}^{t}$ denote the relative position and
orientation, respectively; hats indicate demonstration references, and
$\ominus$ denotes the rotation difference expressed as an
axis--angle vector.

The gesture reward \(r_{\mathrm{ges}}\) promotes alignment of the robot’s finger configuration with the demonstrated gesture. We retarget the demonstrated hand pose to the Wuji hand using~\cite{Qin2023AnyTeleopAG}, as shown in Fig.~\ref{fig:contact_retarget}, and then represent the gesture as a set of key vectors extending from the palm center to the finger joints.
For finger $f$ and joint $j$, the vector is defined as
\begin{equation}
\mathbf{v}^{t}_{f,j}
=
(\mathbf{R}^{H}_t)^{\top}
\left(
\mathbf{p}^{t}_{f,j}-\mathbf{p}^{H}_t
\right),
\end{equation}
where $(\mathbf{p}^{H}_t,\mathbf{R}^{H}_t)$ denotes the current palm pose.
Given the corresponding reference vector $\mathbf{v}^{*}_{f,j}$, we formulate the gesture reward as
\begin{equation}
r_{\mathrm{ges}}^{t}
=
\exp\left(
-\beta
\sum_{f,j}
\left\|
\mathbf{v}^{t}_{f,j}
-
\mathbf{v}^{*}_{f,j}
\right\|_2
\right),
\end{equation}
where $\beta$ controls the sensitivity to gesture deviation. 
This reward assigns higher values to hand configurations that more closely match the reference interaction gesture. Finally, $r_{\mathrm{track}}$ encourages tracking of target positions by awarding a bonus upon reaching each target.
\paragraph{Training Strategy}
\begin{figure}[!h]
    \centering
    \includegraphics[width=\linewidth]{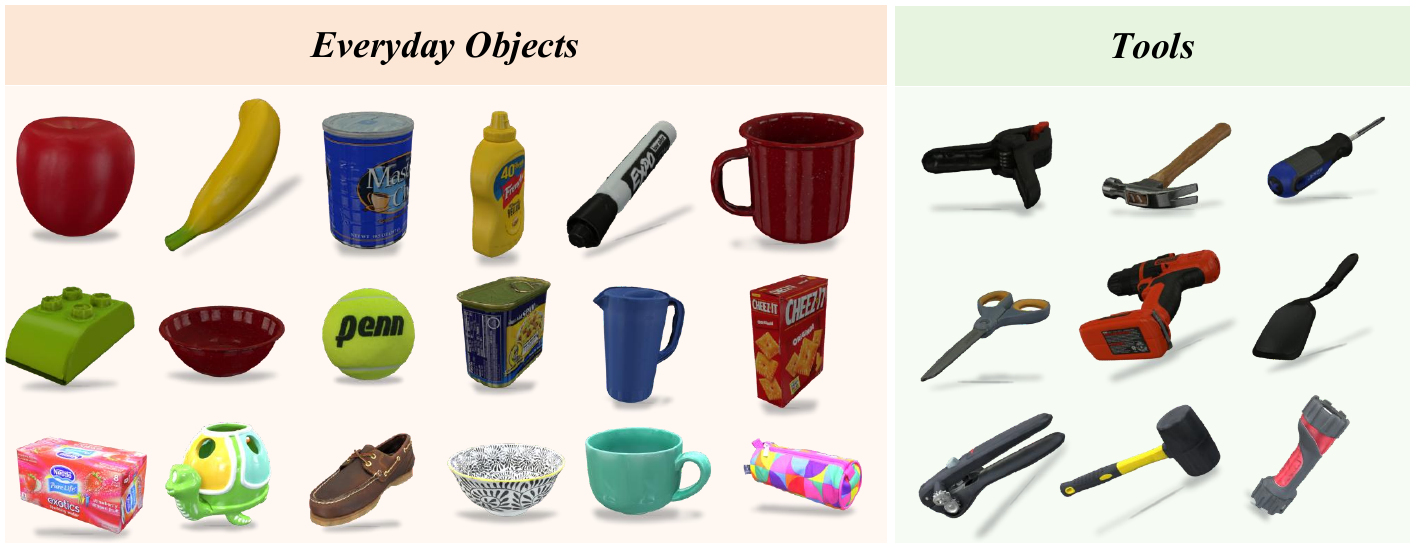}
    \caption{Visualization of sampled objects for policy pretraining. The collection spans diverse categories, including everyday objects and tools, with varied geometries.}
    \label{fig:pretrain_objects}
    \vspace{-1em}
\end{figure}
The high-dimensional action space of the robotic hand 
makes direct policy optimization challenging. We therefore adopt a two-stage training scheme consisting of pretraining and finetuning. The pretraining stage equips the policy with general manipulation capabilities and provides a favorable initialization for downstream learning by training on over 3200 objects~\cite{chao2021dexycb,Xu2023UniDexGraspUR}, as shown in Fig.~\ref{fig:pretrain_objects}. Each object is associated with hundreds of reference hand poses. We then finetune the policy on each task to acquire the desired skills, guided by the object and hand–object interaction
priors extracted from the corresponding HOI video. Notably, pretraining and finetuning share the same policy architecture and reward formulation. We pretrain the policy for 30K iterations and finetune for 10K per task across 4,096 parallel environments. 
SAPG~\cite{singla2024sapg} is used to optimize both stages, with its four blocks employing distinct exploration distributions to collect diverse rollouts and broaden coverage of the action space.
\subsubsection{DAgger Distillation Training}
Given the optimized teacher policies after finetuning, we distill them into deployable student policies via DAgger \cite{ross2011dagger}, enabling real-world execution under partial observations. The student policy adopts the same actor architecture as the teacher. It excludes finger–object
distances and contact information and is trained to match the teacher’s actions by minimizing the objective:
\begin{equation}
\mathcal{L}_{\mathrm{distill}}
=
\left\|
\pi^{\mathrm{S}}\!\left(o_t^{\mathrm{partial}}\right)
-
\pi^{\mathrm{T}}\!\left(o_t^{\mathrm{priv}}\right)
\right\|_2^2 ,
\end{equation}
where $o_t^{\mathrm{priv}}$ and $o_t^{\mathrm{partial}}$ denote the privileged and real-world-accessible observations, respectively.

\section{Experiments}
\label{sec:experiments}
\subsection{Experimental Setup}
We benchmark our method against two object-centric approaches, SimToolReal and Dex4D, across eight manipulation tasks. For each baseline, we follow its original training protocol and introduce an additional finetuning stage for task-specific adaptation. The tasks span diverse hand–object interactions, including tool use, object transport, placement, and stacking. We briefly describe each task below.
\begin{itemize}
    \item \textbf{Nail Hammering}: Grasp a hammer to strike a nail.
    \item \textbf{Umbrella
Hanging}: Hang an umbrella on a target hook.
    \item \textbf{Debris
Sweeping}: Sweep debris into a dustpan.
    \item \textbf{Pepper Lifting}: Stably lift a pepper from the tabletop.
    \item \textbf{Eggplant
Placement}: Place an eggplant in a basket.
    \item \textbf{Bowl Stacking}: Stack one bowl on top of another.
    \item \textbf{Cup
Relocation}: Place a cup on the wooden rack.
    \item \textbf{Upright
Sprayer}: Reorient a spray bottle upright.
\end{itemize}

\begin{table*}[!t]
  \centering
  \caption{Success rates (\%) across eight manipulation tasks in simulation, with averages in the final column.}
  \label{tab:success_tasks}
  \small
  \setlength{\tabcolsep}{5.5pt}
  \renewcommand{\arraystretch}{1.15}

  \resizebox{0.92\textwidth}{!}{%
  \begin{tabular}{lccccccccc}
    \toprule
    Policy &
    \makecell{Nail\\Hammering} &
    \makecell{Umbrella\\Hanging} &
    \makecell{Debris\\Sweeping} &
    \makecell{Upright\\Sprayer} &
    \makecell{Cup\\Relocation} &
    \makecell{Bowl\\Stacking} &
    \makecell{Eggplant\\Placement} &
    \makecell{Pepper\\Lifting} &
    Avg. \\
    \midrule

    SimToolReal
    & 39.2 & 48.2 & 42.3 & 64.6
    & 64.4 & 69.0 & 77.1 & 53.6
    & 57.3 \\

    Dex4D
    & 57.8 & 42.7 & 75.3 & 78.4
    & 69.9 & 58.0 & 89.6 & 62.5
    & 66.8 \\

    \textbf{HandMimicX}
    & \textbf{90.5}
    & \textbf{83.9}
    & \textbf{85.7}
    & \textbf{81.9}
    & \textbf{84.9}
    & \textbf{75.2}
    & \textbf{92.9}
    & \textbf{87.8}
    & \textbf{85.4} \\

    \bottomrule
  \end{tabular}%
  \vspace{-1em}
  }

\end{table*}

\begin{figure}[!h]
    \centering
    \includegraphics[width=\linewidth]{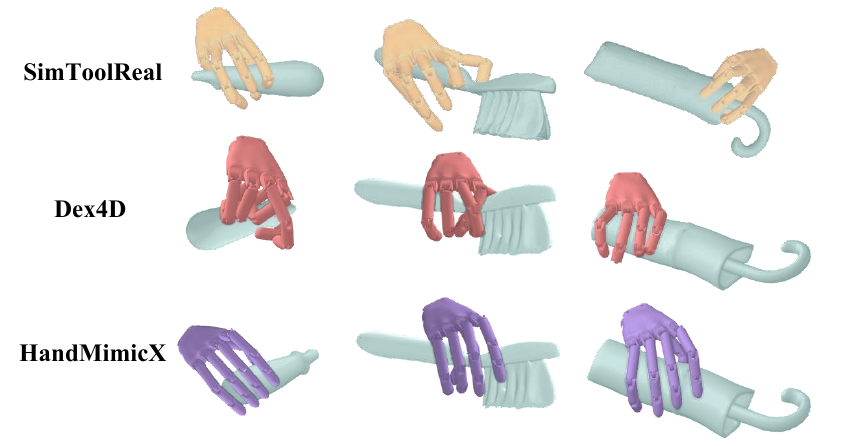}
    \caption{Visualization of hand–object interaction poses from SimToolReal, Dex4D, and HandMimicX. Our HandMimicX produces the most anatomically natural hand poses.}
    \label{fig:handpose}
    \vspace{-1em}
\end{figure}
\subsection{Experiments in Simulation}
 We compare our method with the baselines in simulation by evaluating their ability to move objects through a sequence of randomly sampled waypoints, following the evaluation protocol of Dex4D and SimToolReal. The target advances to the next waypoint once the current waypoint is reached, and a trial is considered successful if all 5 waypoints are reached. Success rates are computed across 4,096 environments. As reported in Tab.~\ref{tab:success_tasks}, our method achieves substantially higher success rates than the baselines, with an average success rate of 85.4\%.
\paragraph{Visualization of Hand-Object Interaction Poses} 
In addition to improving object manipulation success rates, our interaction formulation plays a more fundamental role in maintaining natural, human-like hand–object configurations. As illustrated in Fig.~\ref{fig:handpose}, our method suppresses implausible hand postures, yielding more realistic finger articulation and appropriate contact with the object.


\paragraph{Ablations} We conduct comprehensive ablation studies to systematically assess the contributions of point correspondence and the interaction prior. Specifically, we separately remove the feature fusion module from the correspondence-aware encoder and the interaction prior regularization to form two ablated variants, while keeping all other training settings unchanged. These ablations isolate the role of point correspondence in encoding the relative geometry between the current and target object configurations, as well as the effect of the interaction prior on maintaining plausible and desired hand poses. Tab.~\ref{tab:component_ablation} shows that both components improve performance individually and yield further gains when combined.
To further assess the contribution of pretraining, we conduct an additional ablation study comparing policies trained with and without pretraining. The results show that pretraining increases the average success rate by 12.1\%, suggesting that it provides a more effective initialization for subsequent policy optimization.

\begin{table*}[!t]
  \centering
  \caption{Ablation study evaluating the contributions of the interaction-pose prior, point-correspondence module, and policy pretraining to the overall manipulation performance of our framework. Results are reported as task success rates (\%) across different ablation settings, with averages in the final column.}
  \label{tab:component_ablation}
  \small
  \setlength{\tabcolsep}{5.5pt}
  \renewcommand{\arraystretch}{1.15}

  \resizebox{0.92\textwidth}{!}{%
  \begin{tabular}{lccccccccc}
    \toprule
    Policy &
    \makecell{Nail\\Hammering} &
    \makecell{Umbrella\\Hanging} &
    \makecell{Debris\\Sweeping} &
    \makecell{Upright\\Sprayer} &
    \makecell{Cup\\Relocation} &
    \makecell{Bowl\\Stacking} &
    \makecell{Eggplant\\Placement} &
    \makecell{Pepper\\Lifting} &
    Avg. \\
    \midrule

    w/o Pose Prior
    & 78.3 & 64.1 & 79.4 & 64.5
    & 69.7 & 76.6 & 92.5 & 75.9
    & 75.1 \\

    w/o Correspondence
    & 85.8 & 81.7 & 82.3 & 72.1
    & 78.5 & 64.4 & 87.4 & 74.2
    & 78.3 \\

    w/o Pretraining
    & 79.7 & 71.6 & 62.2 & 71.3
    & 75.0 & 61.8 & 90.6 & 74.1
    & 73.3 \\

    \textbf{HandMimicX}
    & \textbf{90.5}
    & \textbf{83.9}
    & \textbf{85.7}
    & \textbf{81.9}
    & \textbf{84.9}
    & \textbf{75.2}
    & \textbf{92.9}
    & \textbf{87.8}
    & \textbf{85.4} \\

    \bottomrule
  \end{tabular}%
  }

\end{table*}
\begin{table}[!h]
  \centering
  \caption{Skill transfer evaluation across eight real-world manipulation tasks. Each cell reports successful trials / total trials. The final column shows the average success rate.}
  \label{tab:real_robot}
  \small
  \setlength{\tabcolsep}{5.5pt}
  \renewcommand{\arraystretch}{1.15}

  \resizebox{\columnwidth}{!}{%
  \begin{tabular}{ccccccccc}
    \toprule
    \makecell{Nail\\Hammering} &
    \makecell{Umbrella\\Hanging} &
    \makecell{Debris\\Sweeping} &
    \makecell{Upright\\Sprayer} &
    \makecell{Cup\\Relocation} &
    \makecell{Bowl\\Stacking} &
    \makecell{Eggplant\\Placement} &
    \makecell{Pepper\\Lifting} &
    Avg. (\%) \\
    \midrule
{\large 8/10} &
{\large 7/10} &
{\large 7/10} &
{\large 6/10} &
{\large 7/10} &
{\large 6/10} &
{\large 8/10} &
{\large 8/10} &
{\large 71.3} \\
    \bottomrule
  \end{tabular}%
  }
  \vspace{-0.8em}
\end{table}
\begin{figure*}[!h]
    \centering
    \includegraphics[width=\textwidth]{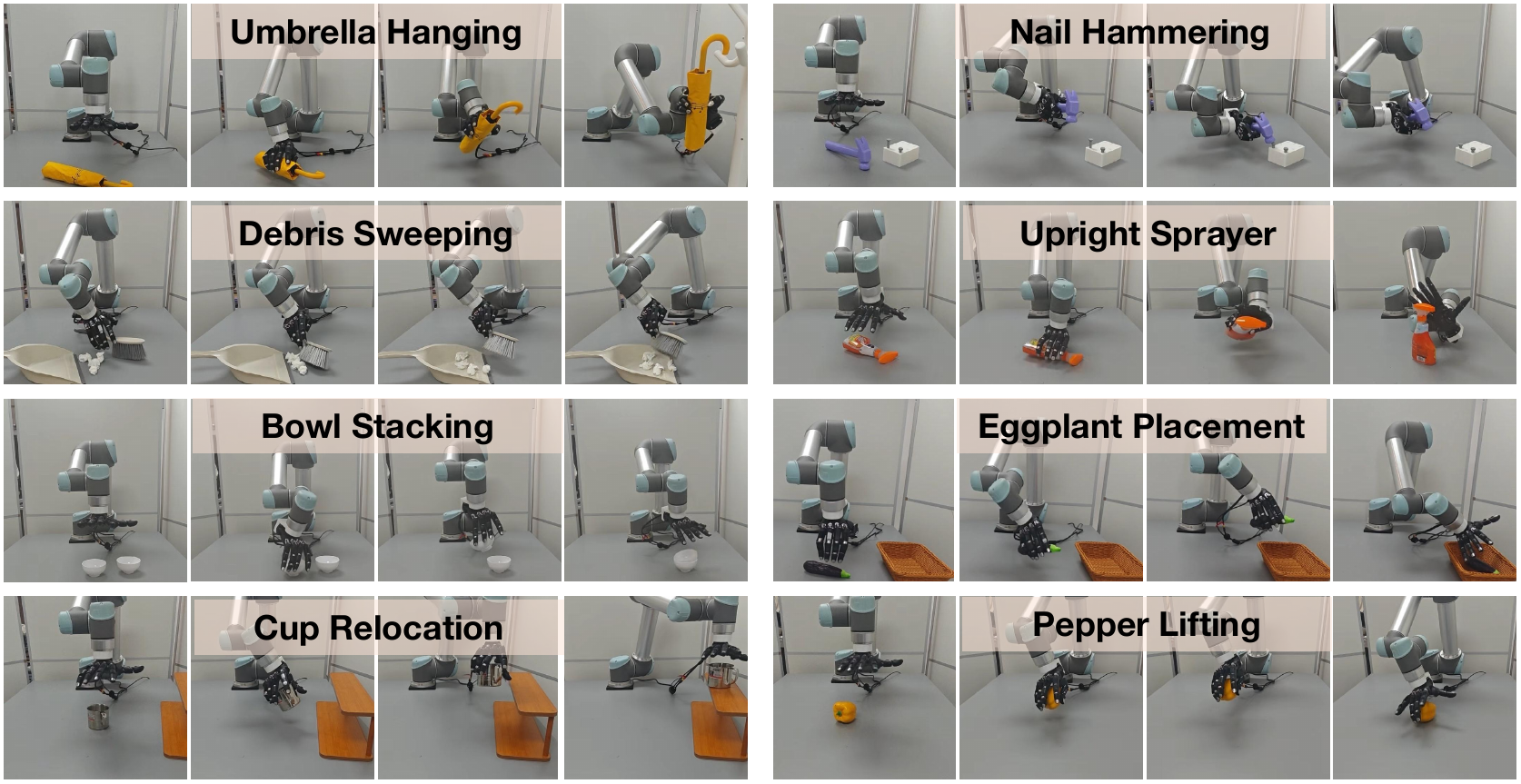}
    \caption{Visualization of real-world execution with HandMimicX. Object trajectories extracted from human demonstration videos serve as targets for robot execution (please refer to the accompanying video for further visualizations).}
    \label{fig:real_world}
\end{figure*}
\begin{figure}[!h]
    \centering
    \includegraphics[width=\linewidth]{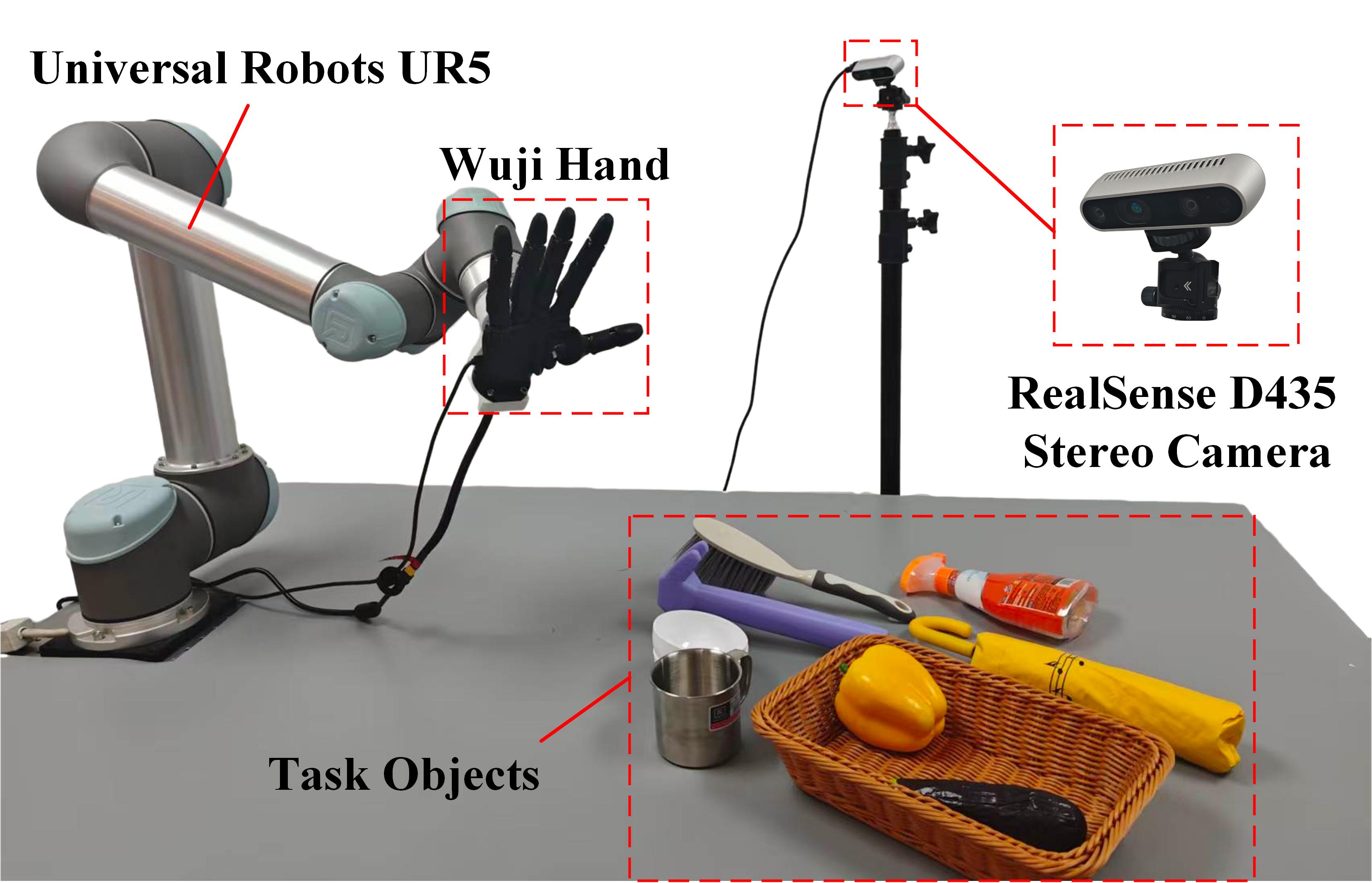}
    \caption{Real-world hardware setup and objects used in our experiments. The system comprises a 6-DoF UR5 robotic arm equipped with a 20-DoF Wuji hand and a single side-mounted Intel RealSense D435 stereo camera.}
    \label{fig:hardware_setup}
    \vspace{-1.5em}
\end{figure}
\subsection{Real-World Deployment}

As shown in Fig.~\ref{fig:hardware_setup}, our system comprises a 6-DoF UR5 robotic arm equipped with a 20-DoF Wuji hand. A single side-mounted Intel RealSense D435 camera provides visual observations for object tracking. We calibrate the camera extrinsics with respect to the robot base frame. During inference, 
FoundationPose estimates the object’s 6-DoF pose. Points sampled from the object mesh are transformed into the robot base frame using calibrated camera extrinsics to obtain the current object point clouds, synchronized with robot proprioception. The goal point cloud is predefined according to the manipulation objective. The student policy is conditioned on these inputs to predict the next action.

\begin{figure}[!h]
    \centering
    \includegraphics[width=\linewidth,trim={0 0 0 0mm},clip]{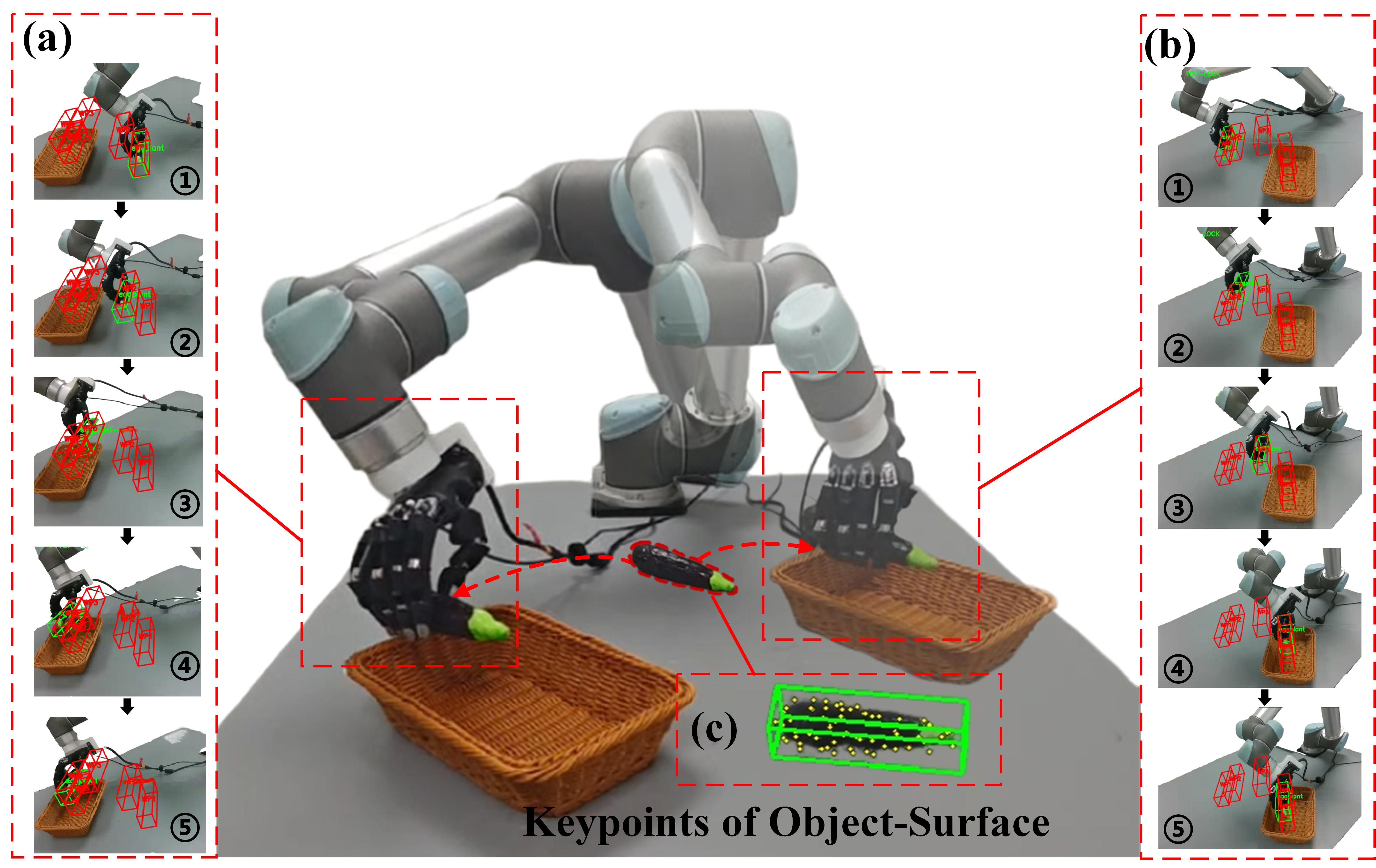}
    \caption{Skill transfer from a single object trajectory extracted from egocentric video.  Panels (a) and (b) show executions from different initial object positions and orientations. Panel (c) visualizes the 128 canonical surface keypoints used to represent the object state for the actor.}
\label{fig:execution_process_one2many}
\vspace{-0.8em}
\end{figure}

Our real-world evaluation examines whether the proposed object-centric policy captures the manipulation skills underlying human video demonstrations and generalizes their execution across spatially varying trajectories. For each dexterous manipulation task, we conduct 10 trials, each of which requires the policy to execute a distinct object trajectory derived from the demonstrated motion. Specifically, we apply affine transformations to the demonstrated trajectory to vary its intermediate waypoints and target object pose. Fig.~\ref{fig:real_world} shows real-world executions of the object trajectories demonstrated in the videos for all eight tasks, while Fig.~\ref{fig:execution_process_one2many} illustrates spatial generalization through the execution of diverse trajectories.
The quantitative results in Tab.~\ref{tab:real_robot} substantiate the effective transfer of the learned generalizable manipulation skills to the physical system.
\section{Conclusion}
\label{sec:conclusion}
We presented HandMimicX, an object-centric reinforcement learning framework for acquiring a dexterous manipulation skill from a single egocentric human demonstration. Our framework integrates depth-refined hand--object reconstruction, correspondence-aware object representations, and interaction priors that regularize contact formation and finger configurations. By coupling object-level goals with demonstrated interaction structure, HandMimicX enables the robot to adapt its manipulation behavior to spatial configurations beyond the original demonstration. Experiments across eight tasks show improved success rates over object-centric baselines, while ablations support the contributions of interaction priors, geometric correspondence, and pretraining. Real-world deployment also demonstrates the feasibility of transferring these learned behaviors to physical manipulation.

\section{Limitations}
Despite its strong performance, our framework has several limitations. It relies on reconstructed object geometry, estimated hand poses, and contact predictions, whose errors can propagate into policy learning. Moreover, our contact representation is limited to discrete region-level states, without modeling finer-grained physical contact forces. Our evaluation of the framework across different embodiments and on bimanual manipulation tasks also remains limited in scope. Future work will investigate more robust human–object interaction perception pipelines, incorporate force-aware contact representations, and extend the framework to more embodiments and manipulation tasks.


\FloatBarrier
\bibliographystyle{IEEEtran}
\bibliography{main}

@inproceedings{chen2026sam3d,
  title={{SAM 3D}: 3Dfy Anything in Images},
  author={Chen, Xingyu and Chu, Fu-Jen and Gleize, Pierre and Liang, Kevin J. and Sax, Alexander and Tang, Hao and Wang, Weiyao and Guo, Michelle and Hardin, Thibaut and Li, Xiang and Lin, Aohan and Liu, Jia-Wei and Ma, Ziqi and Sagar, Anushka and Song, Bowen and Wang, Xiaodong and Yang, Jianing and Zhang, Bowen and Doll{\'a}r, Piotr and Gkioxari, Georgia and Feiszli, Matt and Malik, Jitendra},
  booktitle={Proceedings of the IEEE/CVF Conference on Computer Vision and Pattern Recognition (CVPR)},
  pages={7220--7232},
  year={2026}
}

@inproceedings{jung2025haco,
  title={Learning Dense Hand Contact Estimation from Imbalanced Data},
  author={Jung, Daniel Sungho and Lee, Kyoung Mu},
  booktitle={Advances in Neural Information Processing Systems},
  volume={38},
  year={2025},
  doi={10.52202/085713-4017}
}

@inproceedings{ross2011dagger,
  title={A Reduction of Imitation Learning and Structured Prediction to No-Regret Online Learning},
  author={Ross, Stephane and Gordon, Geoffrey and Bagnell, Drew},
  booktitle={Proceedings of the Fourteenth International Conference on Artificial Intelligence and Statistics},
  pages={627--635},
  volume={15},
  series={Proceedings of Machine Learning Research},
  publisher={PMLR},
  year={2011}
}

@inproceedings{pavlakos2024reconstructing,
  title={Reconstructing hands in 3d with transformers},
  author={Pavlakos, Georgios and Shan, Dandan and Radosavovic, Ilija and Kanazawa, Angjoo and Fouhey, David and Malik, Jitendra},
  booktitle={2024 IEEE/CVF Conference on Computer Vision and Pattern Recognition (CVPR)},
  pages={9826--9836},
  year={2024},
  organization={IEEE}
}

@inproceedings{yu2025dyn,
  title={Dyn-hamr: Recovering 4d interacting hand motion from a dynamic camera},
  author={Yu, Zhengdi and Zafeiriou, Stefanos and Birdal, Tolga},
  booktitle={2025 IEEE/CVF Conference on Computer Vision and Pattern Recognition (CVPR)},
  pages={27716--27726},
  year={2025},
  organization={IEEE}
}

@inproceedings{zhang2025hawor,
  title={Hawor: World-space hand motion reconstruction from egocentric videos},
  author={Zhang, Jinglei and Deng, Jiankang and Ma, Chao and Potamias, Rolandos Alexandros},
  booktitle={2025 IEEE/CVF Conference on Computer Vision and Pattern Recognition (CVPR)},
  pages={1805--1815},
  year={2025},
  organization={IEEE}
}

@inproceedings{fan2024hold,
  title={Hold: Category-agnostic 3d reconstruction of interacting hands and objects from video},
  author={Fan, Zicong and Parelli, Maria and Kadoglou, Maria Eleni and Chen, Xu and Kocabas, Muhammed and Black, Michael J and Hilliges, Otmar},
  booktitle={2024 IEEE/CVF Conference on Computer Vision and Pattern Recognition (CVPR)},
  pages={494--504},
  year={2024},
  organization={IEEE}
}

@inproceedings{qu2023novel,
  title={Novel-view synthesis and pose estimation for hand-object interaction from sparse views},
  author={Qu, Wentian and Cui, Zhaopeng and Zhang, Yinda and Meng, Chenyu and Ma, Cuixia and Deng, Xiaoming and Wang, Hongan},
  booktitle={2023 IEEE/CVF International Conference on Computer Vision (ICCV)},
  pages={15054--15065},
  year={2023},
  organization={IEEE}
}

@inproceedings{zhang2026robowheel,
  title={Robowheel: A data engine from real-world human demonstrations for cross-embodiment robotic learning},
  author={Zhang, Yuhong and Gao, Zihan and Li, Shengpeng and Chen, Ling-Hao and Liu, Kaisheng and Cheng, Runqing and Lin, Xiao and Liu, Junjia and Li, Zhuoheng and Feng, Jingyi and others},
  booktitle={Proceedings of the IEEE/CVF Conference on Computer Vision and Pattern Recognition},
  pages={6664--6674},
  year={2026}
}

@article{chen2026dexterous,
  title={Dexterous manipulation policies from rgb human videos via 3d hand-object trajectory reconstruction},
  author={Chen, Hongyi and Dong, Tony and Wu, Tiancheng and Wang, Liquan and Jangir, Yash and Niu, Yaru and Ye, Yufei and Bharadhwaj, Homanga and Erickson, Zackory and Ichnowski, Jeffrey},
  journal={arXiv preprint arXiv:2602.09013},
  year={2026}
}

@article{ren2026c2dex,
  title={C2Dex: Contact-Consistent Reconstruction and Retargeting for Dexterous Manipulation from Monocular Video},
  author={Ren, Jie and Jiang, Zhehao and Yang, Yinhong and Jia, Haorui and Jiang, Han and Li, Ben and Yao, Yao and Lin, Cheng and Shen, Qiu and Bing, Zhenshan and others},
  journal={arXiv preprint arXiv:2608.07045},
  year={2026}
}

@article{li2025scalable,
  title={Scalable vision-language-action model pretraining for robotic manipulation with real-life human activity videos},
  author={Li, Qixiu and Deng, Yu and Liang, Yaobo and Luo, Lin and Zhou, Lei and Yao, Chengtang and Zeng, Lingqi and Feng, Zhiyuan and Liang, Huizhi and Xu, Sicheng and others},
  journal={arXiv preprint arXiv:2510.21571},
  year={2025}
}

@article{zheng2026egoscale,
  title={Egoscale: Scaling dexterous manipulation with diverse egocentric human data},
  author={Zheng, Ruijie and Niu, Dantong and Xie, Yuqi and Wang, Jing and Xu, Mengda and Jiang, Yunfan and Casta{\~n}eda, Fernando and Hu, Fengyuan and Tan, You Liang and Fu, Letian and others},
  journal={arXiv preprint arXiv:2602.16710},
  year={2026}
}

@article{luo2025being,
  title={Being-h0: vision-language-action pretraining from large-scale human videos},
  author={Luo, Hao and Feng, Yicheng and Zhang, Wanpeng and Zheng, Sipeng and Wang, Ye and Yuan, Haoqi and Liu, Jiazheng and Xu, Chaoyi and Jin, Qin and Lu, Zongqing},
  journal={arXiv preprint arXiv:2507.15597},
  year={2025}
}

@inproceedings{bi2026h,
  title={H-rdt: Human manipulation enhanced bimanual robotic manipulation},
  author={Bi, Hongzhe and Wu, Lingxuan and Lin, Tianwei and Tan, Hengkai and Su, Zhizhong and Su, Hang and Zhu, Jun},
  booktitle={Proceedings of the AAAI Conference on Artificial Intelligence},
  volume={40},
  number={22},
  pages={18135--18143},
  year={2026}
}

@article{makoviychuk2021isaac,
  title={Isaac gym: High performance gpu-based physics simulation for robot learning},
  author={Makoviychuk, Viktor and Wawrzyniak, Lukasz and Guo, Yunrong and Lu, Michelle and Storey, Kier and Macklin, Miles and Hoeller, David and Rudin, Nikita and Allshire, Arthur and Handa, Ankur and others},
  journal={arXiv preprint arXiv:2108.10470},
  year={2021}
}

@inproceedings{todorov2012mujoco,
  title={Mujoco: A physics engine for model-based control},
  author={Todorov, Emanuel and Erez, Tom and Tassa, Yuval},
  booktitle={2012 IEEE/RSJ international conference on intelligent robots and systems},
  pages={5026--5033},
  year={2012},
  organization={IEEE}
}

@inproceedings{qin2022dexmv,
  title={Dexmv: Imitation learning for dexterous manipulation from human videos},
  author={Qin, Yuzhe and Wu, Yueh-Hua and Liu, Shaowei and Jiang, Hanwen and Yang, Ruihan and Fu, Yang and Wang, Xiaolong},
  booktitle={European Conference on Computer Vision},
  pages={570--587},
  year={2022},
  organization={Springer}
}

@inproceedings{mandikal2022dexvip,
  title={Dexvip: Learning dexterous grasping with human hand pose priors from video},
  author={Mandikal, Priyanka and Grauman, Kristen},
  booktitle={Conference on Robot Learning},
  pages={651--661},
  year={2022},
  organization={PMLR}
}

@inproceedings{hoque2026egodex,
  title={Egodex: Learning dexterous manipulation from large-scale egocentric video},
  author={Hoque, Ryan and Huang, Peide and Yoon, David and Zhang, Jian and others},
  booktitle={International Conference on Learning Representations},
  volume={2026},
  pages={4218--4237},
  year={2026}
}

@inproceedings{chao2021dexycb,
  title={DexYCB: A Benchmark for Capturing Hand Grasping of Objects},
  author={Chao, Yu-Wei and Yang, Wei and Xiang, Yu and Molchanov, Pavlo and Handa, Ankur and Tremblay, Jonathan and Narang, Yashraj S and Van Wyk, Karl and Iqbal, Umar and Birchfield, Stan and others},
  booktitle={2021 IEEE/CVF Conference on Computer Vision and Pattern Recognition (CVPR)},
  pages={9040--9049},
  year={2021},
  organization={IEEE}
}

@inproceedings{banerjee2025hot3d,
  title={Hot3d: Hand and object tracking in 3d from egocentric multi-view videos},
  author={Banerjee, Prithviraj and Shkodrani, Sindi and Moulon, Pierre and Hampali, Shreyas and Han, Shangchen and Zhang, Fan and Zhang, Linguang and Fountain, Jade and Miller, Edward and Basol, Selen and others},
  booktitle={2025 IEEE/CVF Conference on Computer Vision and Pattern Recognition (CVPR)},
  pages={7061--7071},
  year={2025},
  organization={IEEE}
}

@inproceedings{zhan2024oakink2,
  title={Oakink2: A dataset of bimanual hands-object manipulation in complex task completion},
  author={Zhan, Xinyu and Yang, Lixin and Zhao, Yifei and Mao, Kangrui and Xu, Hanlin and Lin, Zenan and Li, Kailin and Lu, Cewu},
  booktitle={2024 IEEE/CVF Conference on Computer Vision and Pattern Recognition (CVPR)},
  pages={445--456},
  year={2024},
  organization={IEEE}
}

@article{Kwon2021H2OTH,
  title={H2O: Two Hands Manipulating Objects for First Person Interaction Recognition},
  author={Taein Kwon and Bugra Tekin and Jan St{\"u}hmer and Federica Bogo and Marc Pollefeys},
  journal={2021 IEEE/CVF International Conference on Computer Vision (ICCV)},
  year={2021},
  pages={10118-10128},
  url={https://api.semanticscholar.org/CorpusID:233346960}
}

@article{Chen2024ViViDexLV,
  title={ViViDex: Learning Vision-Based Dexterous Manipulation from Human Videos},
  author={Zerui Chen and Shizhe Chen and Etienne Arlaud and Ivan Laptev and Cordelia Schmid},
  journal={2025 IEEE International Conference on Robotics and Automation (ICRA)},
  year={2024},
  pages={3336-3343},
  url={https://api.semanticscholar.org/CorpusID:269330215}
}

@article{Xu2026DemoBotEL,
  title={DemoBot: Efficient Learning of Bimanual Manipulation with Dexterous Hands From Third-Person Human Videos},
  author={Yucheng Xu and Xiaofeng Mao and Elle Miller and Xinyu Yi and Yang Li and Zhibin Li and Robert B. Fisher},
  journal={ArXiv},
  year={2026},
  volume={abs/2601.01651},
}

@article{Ye2025Gen2RealTD,
  title={Gen2Real: Towards Demo-Free Dexterous Manipulation by Harnessing Generated Video},
  author={Kai Ye and Yuhang Wu and Shuyuan Hu and Junliang Li and Menghong Liu and Yongquan Chen and Rui Huang},
  journal={ArXiv},
  year={2025},
  volume={abs/2509.14178}
}

@article{Zhou2023LearningHA,
  title={Learning Hybrid Actor-Critic Maps for 6D Non-Prehensile Manipulation},
  author={Wen-Min Zhou and Bowen Jiang and Fan Yang and Chris Paxton and David Held},
  journal={ArXiv},
  year={2023},
  volume={abs/2305.03942},
}

@inproceedings{Qin2022DexPointGP,
  title={DexPoint: Generalizable Point Cloud Reinforcement Learning for Sim-to-Real Dexterous Manipulation},
  author={Yuzhe Qin and Binghao Huang and Zhao-Heng Yin and Hao Su and Xiaolong Wang},
  booktitle={Conference on Robot Learning},
  year={2022},
}

@article{Dan2025XSimCL,
  title={X-Sim: Cross-Embodiment Learning via Real-to-Sim-to-Real},
  author={Prithwish Dan and K. Kedia and Angela Chao and Edward Weiyi Duan and Maximus Adrian Pace and Wei-Chiu Ma and Sanjiban Choudhury},
  journal={ArXiv},
  year={2025},
  volume={abs/2505.07096},
  url={https://api.semanticscholar.org/CorpusID:278501460}
}

@article{Wen2023FoundationPoseU6,
  title={FoundationPose: Unified 6D Pose Estimation and Tracking of Novel Objects},
  author={Bowen Wen and Wei Yang and Jan Kautz and Stanley T. Birchfield},
  journal={2024 IEEE/CVF Conference on Computer Vision and Pattern Recognition (CVPR)},
  year={2023},
  pages={17868-17879},
  url={https://api.semanticscholar.org/CorpusID:266191252}
}

@article{Qin2023AnyTeleopAG,
  title={AnyTeleop: A General Vision-Based Dexterous Robot Arm-Hand Teleoperation System},
  author={Yuzhe Qin and Wei Yang and Binghao Huang and Karl Van Wyk and Hao Su and Xiaolong Wang and Yu-Wei Chao and Dieter Fox},
  journal={ArXiv},
  year={2023},
  volume={abs/2307.04577},
  url={https://api.semanticscholar.org/CorpusID:259367735}
}

@article{Qi2016PointNetDL,
  title={PointNet: Deep Learning on Point Sets for 3D Classification and Segmentation},
  author={C. Qi and Hao Su and Kaichun Mo and Leonidas J. Guibas},
  journal={2017 IEEE Conference on Computer Vision and Pattern Recognition (CVPR)},
  year={2016},
  pages={77-85},
  url={https://api.semanticscholar.org/CorpusID:5115938}
}

@article{Lum2025CrossingTH,
  title={Crossing the Human-Robot Embodiment Gap with Sim-to-Real RL using One Human Demonstration},
  author={Tyler Ga Wei Lum and Olivia Y. Lee and C. Karen Liu and Jeannette Bohg},
  journal={ArXiv},
  year={2025},
  volume={abs/2504.12609},
  url={https://api.semanticscholar.org/CorpusID:277857482}
}

@article{Kuang2026Dex4DTP,
  title={Dex4D: Task-Agnostic Point Track Policy for Sim-to-Real Dexterous Manipulation},
  author={Yuxuan Kuang and Sungjae Park and Katerina Fragkiadaki and Shubham Tulsiani},
  journal={ArXiv},
  year={2026},
  volume={abs/2602.15828},
  url={https://api.semanticscholar.org/CorpusID:285659883}
}

@article{Li2025ManipTransED,
  title={ManipTrans: Efficient Dexterous Bimanual Manipulation Transfer via Residual Learning},
  author={Kailin Li and Puhao Li and Tengyu Liu and Yuyang Li and Siyuan Huang},
  journal={2025 IEEE/CVF Conference on Computer Vision and Pattern Recognition (CVPR)},
  year={2025},
  pages={6991-7003},
  url={https://api.semanticscholar.org/CorpusID:277435197}
}

@article{Xu2023UniDexGraspUR,
  title={UniDexGrasp: Universal Robotic Dexterous Grasping via Learning Diverse Proposal Generation and Goal-Conditioned Policy},
  author={Yinzhen Xu and Weikang Wan and Jialiang Zhang and Haoran Liu and Zikang Shan and Hao Shen and Ruicheng Wang and Haoran Geng and Yijia Weng and Jiayi Chen and Tengyu Liu and Li Yi and He Wang},
  journal={2023 IEEE/CVF Conference on Computer Vision and Pattern Recognition (CVPR)},
  year={2023},
  pages={4737-4746},
  url={https://api.semanticscholar.org/CorpusID:257280140}
}

@article{Yuan2025HERMESHE,
  title={HERMES: Human-to-Robot Embodied Learning from Multi-Source Motion Data for Mobile Dexterous Manipulation},
  author={Zhecheng Yuan and Tianming Wei and Langzhe Gu and Pu Hua and Tianhai Liang and Yuanpei Chen and Huazhe Xu},
  journal={ArXiv},
  year={2025},
  volume={abs/2508.20085},
  url={https://api.semanticscholar.org/CorpusID:280919277}
}

@article{Kedia2026SimToolRealAO,
  title={SimToolReal: An Object-Centric Policy for Zero-Shot Dexterous Tool Manipulation},
  author={K. Kedia and Tyler Ga Wei Lum and Jeannette Bohg and C. Karen Liu},
  journal={ArXiv},
  year={2026},
  volume={abs/2602.16863},
  url={https://api.semanticscholar.org/CorpusID:285787326}
}

@article{fu2025cordvip,
  title={Cordvip: Correspondence-based visuomotor policy for dexterous manipulation in real-world},
  author={Fu, Yankai and Feng, Qiuxuan and Chen, Ning and Zhou, Zichen and Liu, Mengzhen and Wu, Mingdong and Chen, Tianxing and Rong, Shanyu and Liu, Jiaming and Dong, Hao and others},
  journal={arXiv preprint arXiv:2502.08449},
  year={2025}
}

@article{Zhang2026UniDexAR,
  title={UniDex: A Robot Foundation Suite for Universal Dexterous Hand Control from Egocentric Human Videos},
  author={Gu Zhang and Qicheng Xu and Haozhe Zhang and Jianhang Ma and Long He and Yiming Bao and Zeyu Ping and Zhecheng Yuan and Chenhao Lu and Chengbo Yuan and Tianhai Liang and Xiaoyu Tian and Maanping Shao and Feihong Zhang and Mingyu Ding and Yang Gao and Hao Zhao and Hang Zhao and Huazhe Xu},
  journal={ArXiv},
  year={2026},
  volume={abs/2603.22264},
  url={https://api.semanticscholar.org/CorpusID:286766691}
}

@inproceedings{singla2024sapg,
title={{SAPG}: Split and Aggregate Policy Gradients},
author={Singla, Jayesh and Agarwal, Ananye and Pathak, Deepak},
booktitle={Proceedings of the 41st International Conference on Machine Learning},
pages={45759--45772},
volume={235},
series={Proceedings of Machine Learning Research},
publisher={PMLR},
year={2024},}

@article{Han2026Video2Sim2RealFA,
  title={Video2Sim2Real: Full-Stack Autonomous Dexterous Skill Acquisition from a Single Human Video},
  author={Yunhai Han and Jianuo Qiu and Linhao Bai and Ziyu Xiao and Zihang Zeng and Yangcen Liu and Zhaodong Yang and Shalin Jain and Wenrui Ma and Jiaqi Fu and Yuqian Zheng and Manisha Natarajan and Muhammad Zubair Irshad and Kenneth Shaw and Matthew Craig Gombolay and Zsolt Kira and Harish chaandar Ravichandar},
  journal={ArXiv},
  year={2026},
  volume={abs/2606.08828},
  url={https://api.semanticscholar.org/CorpusID:289099227}
}

@article{Liu2025DexTrackTG,
  title={DexTrack: Towards Generalizable Neural Tracking Control for Dexterous Manipulation from Human References},
  author={Xueyi Liu and Jianibieke Adalibieke and Qianwei Han and Yuzhe Qin and Li Yi},
  journal={ArXiv},
  year={2025},
  volume={abs/2502.09614},
  url={https://api.semanticscholar.org/CorpusID:276317383}
}

@article{Feng2026AMR,
  title={A Minimalist Retargeting-Guided Reinforcement Learning Recipe for Dexterous Manipulation},
  author={Yunhai Feng and Natalie Leung and Jiaxuan Wang and Lu-Jie Yang and Haozhi Qi and Preston Culbertson},
  journal={ArXiv},
  year={2026},
  volume={abs/2607.11874},
  url={https://api.semanticscholar.org/CorpusID:290143227}
}
\end{document}